\documentclass[lettersize,journal]{IEEEtran}
\usepackage{amsmath,amsfonts}
\usepackage{algorithmic}
\usepackage{algorithm}
\usepackage{array}
\usepackage[caption=false,font=normalsize,labelfont=sf,textfont=sf]{subfig}
\usepackage{textcomp}
\usepackage{stfloats}
\usepackage{url}
\usepackage{wrapfig}
\usepackage{verbatim}
\usepackage{rotating}
\usepackage{booktabs} 
\usepackage{multirow} 
\usepackage{rotating} 
\usepackage{array}
\usepackage{graphicx}
\usepackage{cite}
\usepackage{hyperref}       
\usepackage{url}            
\usepackage[table]{xcolor}
\usepackage{textcomp}
\usepackage{booktabs}       
\usepackage{amsfonts}       
\usepackage{nicefrac}       
\usepackage{microtype}      
\usepackage{lipsum}
\usepackage{fancyhdr}       
\usepackage{graphicx}       
\graphicspath{{media/}}     
\usepackage{marvosym}
\usepackage{orcidlink}
\usepackage{arydshln}
\usepackage{multirow}
\usepackage{wrapfig}
\usepackage{graphicx}
\usepackage{booktabs}
\usepackage{multirow,amsmath}
\usepackage{amsthm}

\usepackage{xcolor}

\newcommand{\update}[1]{{\textcolor{black}{#1}}}

\theoremstyle{definition} 

\begin{document}

\title{MarkPlugger: Generalizable Watermark Framework for Latent Diffusion Models without Retraining}


\author{Guokai Zhang~\dag, Lanjun Wang~\dag,~\IEEEmembership{Member,~IEEE}, Yuting Su, ~\IEEEmembership{Member,~IEEE}, An-An Liu*, ~\IEEEmembership{Senior Member,~IEEE}
\thanks{This work was supported in by the National Natural Science Foundation of China under No. U21B2024, No. 62425307, and No. 62202329.
}
\thanks{Guokai Zhang, Lanjun Wang, Yuting Su, and An-An Liu are with the
Tianjin University, Tianjin 300072, China.}
\thanks{*Corresponding author: An-An Liu (Email:
anan0422@gmail.com). }
\thanks{\dag These authors contributed equally to this
work.}}

\markboth{Journal of \LaTeX\ Class Files,~Vol.~X, No.~X, September~2024}%
{Shell \MakeLowercase{\textit{et al.}}: A Sample Article Using IEEEtran.cls for IEEE Journals}


\maketitle
\begin{abstract}
Today, the family of latent diffusion models (LDMs) has gained prominence for its high quality outputs and scalability. This has also raised security concerns on social media, as malicious users can create and disseminate harmful content. Existing approaches typically involve training specific components or entire generative models to embed a watermark in generated images for traceability and responsibility. However, in the fast-evolving era of AI-generated content (AIGC), the rapid iteration and modification of LDMs makes retraining with watermark models costly. To address the problem, we propose \textit{MarkPlugger}, a generalizable plug-and-play watermark framework without LDM retraining.  
In particular, to reduce the disturbance of the watermark on the semantics of the generated image, \update{we try to identify a watermark representation that is approaching orthogonal to the semantic in latent space, and apply an additive fusion strategy for the watermark and the semantic.} 
Without modifying any components of the LDMs, we embed diverse watermarks in latent space, adapting to the denoising process. Our experimental findings reveal that our method effectively harmonizes image quality and watermark recovery rate. We also have validated that our method is generalized to multiple official versions and modified variants of LDMs, even without retraining the watermark model. Furthermore, it performs robustly under various attacks of different intensities.
\end{abstract}

\begin{IEEEkeywords}
Latent diffusion model, training-free for LDMs, plug-and-play, watermarking, text-to-image
\end{IEEEkeywords}

\section{Introduction}
\label{sec:intro}
\IEEEPARstart{T}{he} family of diffusion models~\cite{HoJA20, Xiong0F023, NicholD21, SahariaCSLWDGLA22} has revolutionized the landscape of generative models with its unparalleled capabilities, thereby elevating the creativity and artistic expression in advertising, design, and other creative industries. In particular, LDM~\cite{RombachBLEO22} stands out as one of the most widely embraced techniques in the text-to-image (T2I)~\cite{10528891} community, due to its remarkable capability and inventiveness. The artistic creations generated by LDMs even surpass human creativity and authenticity identification. However, loosely regulated platforms have enabled some malicious users to exploit LDMs to generate illegal paintings and disseminate them through social media, making it challenging to trace their origins and assign responsibilities. Instances such as celebrity rumors~\cite{li2024global} and AIGC image fraud cases~\cite{10200392} have been brewing due to source confusion. Consequently, official and commercial institutions, such as the White House and OpenAI, are beginning to recognize these risks and are issuing a series of statements describing security measures~\cite{10401723}.


\begin{figure}[tb]
  \centering
\includegraphics[width=0.5\textwidth]{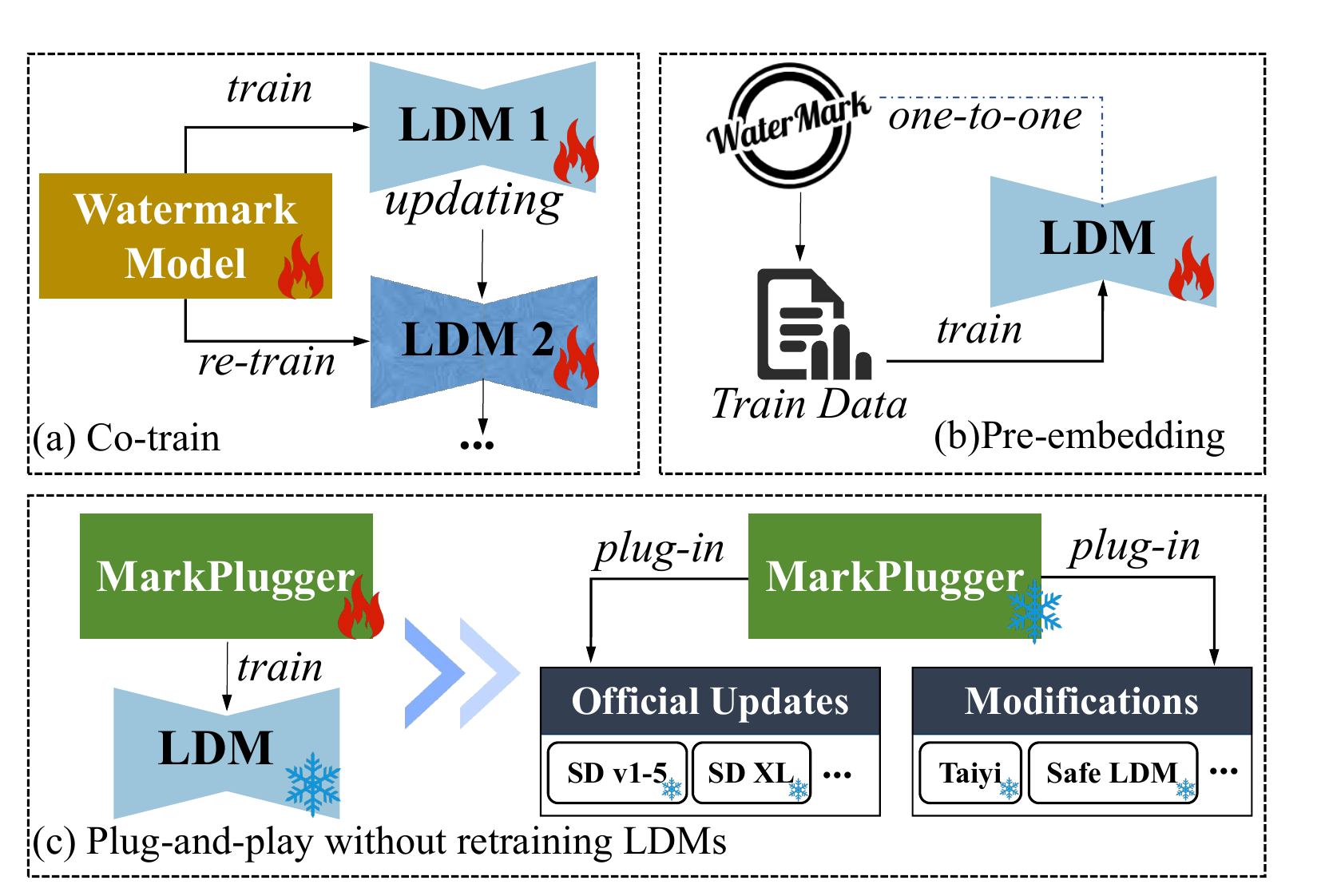}
  \caption{Comparison of our proposed framework with existing methods. One type of method like (a) co-trains the watermark model with the LDM, requiring retraining whenever the model is updating. Another type of method like (b) pre-embeds the watermark into the train data. The one-to-one correspondence between the watermark and the LDM decide the inflexibility of watermark embedding. Therefore, the generalizable plug-and-play watermark without retraining LDMs is proposed, as dipicted in (c).}
  \label{fig:limitations}
  \vspace{-15pt}
\end{figure}

Researchers have acknowledged the inherent risks of image generation methods and have worked to mitigate the potential dangers of misuse related to LDMs. One of the promising solutions is watermarking~\cite{FangJQZZC23}, which improves traceability against the generation of harmful visual content. Two types of methods have been attempted in this respect, though facing certain limitations.
A type of method involves training specific components or entire LDMs with a watermark model to embed watermarks into generated images~\cite{Xiong0F023,Kim_2024_CVPR}, as shown in Fig.~\ref{fig:limitations} (a). However, due to the rapid updates on LDM variants, it is time-consuming and labor-intensive for model holders to take advantage of these training-based methods, leading to poor generalization. Another type of method incorporates a fixed watermark in the training data, allowing the LDM to produce images that retain the watermark~\cite{abs-2308-11123,FernandezCJDF23}, as shown in Fig.~\ref{fig:limitations} (b). Nevertheless, these methods require customizing the watermark data and the watermark model for a specific LDM, which cannot be flexibly deployed. Therefore, to enable the watermark technique to keep up with the advancement of LDMs, it is necessary to consider \textit{how to improve the generalization ability while maintaining the flexibility of the watermark insertion}.  Specifically, improving the generalization ability requires eliminating the need for more extra training on LDMs. Moreover, flexible watermark insertion means simplifying the integration of the watermark model with LDMs without requiring more extensive customization or configuration. It must avoid impairing the denoising process of the latent code.

However, addressing the above research questions encompasses three challenges. First, improving generalization ability relies on reducing the excessive training burden. As the design of watermark models~\cite{Xiong0F023,Kim_2024_CVPR,abs-2308-11123,FernandezCJDF23} is generally dependent on the specific parameters and architecture of an LDM, their co-training involves millions of parameters. Furthermore, once the LDM is modified with various architectures and parameters~\cite{RombachBLEO22, PodellELBDMPR24, fengshenbang}, the watermark method is prone to failure unless it is retrained with LDMs. Even the watermark model requires frequent retraining. Second, image quality is highly susceptible to variations caused by flexible watermark insertion. LDMs heavily depend on the latent code to progressively denoise and facilitate image generation. This dependence determines that subtle perturbations, such as hidden watermark signals in latent code, can be significantly amplified in pixel space and affect the quality of generated images~\cite{ChenGZY023}. Third, ensuring high image quality weakens the watermark signal, reducing its extraction accuracy and robustness. A delicate trade-off is required to minimize the sacrifice to image quality while effectively achieving watermark recoverability. 

In this study, we propose MarkPlugger, a generalizable plug-and-play watermark framework for various versions and variants of LDMs without retraining, which facilitates the T2I task. In the LDM training-free mode, the latent semantic code is deterministic. Our study focuses on exploring the watermark representation that is irrelevant to the semantics in latent space. Specifically, we design a fusion strategy (i.e., additive fusion on a single channel of the latent code). To enhance training efficiency while preserving latent space, we update only the parameters of the watermark encoder-decoder architecture based on the frozen Variational Autoencoder (VAE) from the LDM~\cite{RombachBLEO22}.
Remarkably, our approach remains effective during inference, allowing us to bypass the need for retraining of the LDM. Furthermore, to achieve a balance between image quality and watermark recoverability, we develop an objective function that harmonizes these two factors by regulating both the embedding and extraction of the compressed watermark signal. Extensive experiments validate the superior image quality, watermark invisibility, watermark recoverability, and watermark robustness of our proposed method. We also demonstrate generalization capability across multiple official versions and modified variants of LDMs, including Stable Diffusion (SD) v1-1, v1-4, v1-5, v2-0, v2-1~\cite{RombachBLEO22}, XL~\cite{PodellELBDMPR24}, Taiyi~\cite{fengshenbang} and Safe Latent Diffusion~\cite{SchramowskiBDK23}, all in model training-free mode. 

The main contributions can be summarized as follows:
\begin{itemize}
    \item We propose a generalizable plug-and-play watermark framework for the LDM architecture without retraining, capable of flexibly embedding and extracting diverse watermarks in latent space.
    \item We propose an additive fusion strategy for watermarking in the LDM training-free mode. It focuses on exploring the watermark approximately orthogonal to the semantic.  We have effectively achieved a balance between image quality and watermark recoverability.
    \item Our watermark framework can be generalized across different LDM versions and variants, maintaining consistent performance with almost no degradation.
    \item Extensive experiments show the effectiveness of our proposed approach. Compared with existing methods, our FLOPs are just $1/300$ of the SOTA method, demonstrating their significance as more cost-efficient and user-friendly. 
\end{itemize}

\section{Related Work}
In this section, we begin by highlighting the significant milestones in the development of T2I using diffusion models. Following this, we introduce watermarking techniques for generative tasks, including pre-embedding watermarking, post-embedding watermarking, and joint generation.

\subsection{Diffusion Models}
The family of Denoising Diffusion Probabilistic Models (DDPMs)~\cite{HoJA20,NicholD21,SahariaCSLWDGLA22} has gradually become one of the most important research areas in T2I generation. DDPMs are capable of generating high-quality images from a noise map via multiple iterations. Although eye-catching diffusion models such as DALLE-2~\cite{abs-2204-06125} and Imagen~\cite{SahariaCSLWDGLA22} have demonstrated impressive capacity, their computational inefficiency has become an obstacle to ongoing research within the community and enterprises. To mitigate computational complexity while maintaining high generation capabilities, Rombach et al. \cite{RombachBLEO22} proposed a relatively lightweight architecture, named latent diffusion model. This model compressed the image in the perceptual latent space to facilitate noise adding and removing. After it was trained on the large-scale LAION-5B dataset \cite{SchuhmannBVGWCC22}, Stability AI reported that LDM exhibited improved generation capabilities and was rebranded as Stable Diffusion. Due to its scalability, controllability, and reduced computational expense, SD has been widely improved to generate diverse high-quality images \cite{QinZYFYZWNXSE0X23,10528891}. However, this advancement has simultaneously given rise to ethical dilemmas and legal disputes.

\subsection{Watermarking Generative Models}
Building on the placement of the watermark embedding, we categorize watermarking methods for generative models into three distinct groups: pre-embedding watermarking, which occurs prior to the image generation; co-training, which integrates watermarking concurrently with the image generation but needs to train both LDM and the watermark model; \update{ plug-and-play, which injects watermark with the image generation but without training or fine-tuning LDMs.}

\subsubsection{Pre-embedding watermarking}The pre-embedding watermarking method aims to integrate the single watermark into input training data before generating images. By utilizing the watermarked data to train generative models, the generated images carry the corresponding traceable data, facilitating the tracking of their origin. To address the concerns of facial synthesized images, Zhao et al. \cite{ZhaoLDLZY23} embedded the fixed binary sequence as a watermark into the training dataset. The sequence would be transferred into the generated images at inference time. Building on these approaches, Ditria et al. \cite{abs-2308-11123} showcased that the training with watermarked data was equally effective for SDs. In the scenario of model distribution, Fernandez et al.~\cite{FernandezCJDF23} pre-trained both the watermark encoder and decoder with prepared watermark examples. Subsequently, they integrated the watermark decoder with an LDM to train its VAE decoder, extracting a fixed binary sequence from generated images. These methods have two limitations. First, they require either pre-training of the watermark model or pre-embedding of a fixed watermark into the training data, which becomes costly for large-scale datasets. Second, they can only embed a fixed message as the watermark, unless retraining the model.


\subsubsection{Co-train} The co-train methods combine watermark embedding into the image generation process, considered as a unified procedure~\cite{WangL0F23}. In the scenario of model distribution for the specific user, Kim et al. \cite{Kim_2024_CVPR} embedded a fixed binary sequence into convolutional layers of generative models and conducted training across the entire model. Recently, Wang et al. \cite{WangL0F23} attempted to embed a sequence as a copyright notice at the feature level in the style transfer task for LDM. Given the need for image traceability in the T2I task, Liu et al. \cite{abs-2309-03815} proposed a joint text to watermark and image generation method. This approach leveraged traceability metadata, including user identification, input prompt, and timestamp, to create a binary watermark. For embedding a flexible watermark sequence, Xiong et al. \cite{Xiong0F023} trained the VAE decoder of LDM with the watermark model. Overall, these methods exhibit low risk and low pre-processing costs. Considering the rapid pace of updates in diffusion models, notably LDMs, training watermark models for each variant is impractical. 

{\subsubsection{Plug-and-play} More recently, there are a series of studies injecting watermarks based on the DDIM inversion property~\cite{wen2024tree,yang2024gaussian,zhang2024attack}. 
These studies can also achieve the watermark injection based on a frozen stable diffusion model.  However, despite the fact that they can only work for DDIM, the watermark information injected is sampled from a Gaussian distribution, but cannot embed more flexible and concrete information to record the metadata of the generated image, e.g., the author ID, the prompt, the model version, and generation time.   }Consequently, we propose a generalizable plug-and-play watermark solution without retraining any components of LDMs or relying on the DDIM sampling scheme.

\section{Problem Definition}

In this section, we first outline the preliminaries of generating images using LDMs. Then, we define the plug-and-play watermarking scheme without retraining.

\subsection{Image Generation Using LDM}
The family of LDM represents some of the most widely-used generative models for the T2I task, notable for their high scalability and accessibility to the research community~\cite{RombachBLEO22}. Given the latent code $\mathbf{z}_{t-1} \in \mathbb{R}^{h\times w\times c_1}$ at time $t$, LDMs define the forward diffusion process as a Markov process:
\begin{equation}
q(\mathbf{z}_t | \mathbf{z}_{t-1}) = \mathcal{N}(\mathbf{z}_t; \sqrt{1-\beta_t} \mathbf{z}_{t-1}, \beta_t \mathbf{I})
\end{equation}
which progressively injects the noise to $\mathbf{z}_{t-1}$ with a noise schedule $\beta_t \in (0,1)$. After $T$ times iteration, the $\mathbf{z}_T$ approaches a Gaussian distribution, i.e., \update{$q(\mathbf{z}_T) \sim \mathcal{N}(0,\mathbf{I})$}. The closed-form solution for this sampling is: $\mathbf{z}_t = \sqrt{\bar{\alpha}_t}\mathbf{z}_0 + \sqrt{1-\bar{\alpha}_t}\epsilon_t$, where $\bar{\alpha}_t = \prod\limits_{i=0}^t(1-\beta_t)$. The VAE encoder $\mathrm{\Psi}^*_{e}(\cdot)$ compresses the original image $x$ into a latent representation, to which noise $\epsilon_t$ sampled from $\mathcal{N}(0,\mathbf{I})$ is then added, over $T$ iterations. The forward diffusion process at step $t$ can be illustrated as:
\begin{equation}
\mathbf{z
}_t = F(\epsilon_t,\mathbf{z}_{t-1})
\end{equation}
where $F(\cdot)$ is the noising function. 

\update{
Correspondingly, the reverse diffusion process denoises the Gaussian vector $\mathbf{z'
}_T \sim \mathcal{N}(0,\mathbf{I})$ for each denoising step:
\begin{equation}
p_\theta(\mathbf{z'}_{t-1} | \mathbf{z'}_t, t, \mathbf{c}) = \mathcal{N}(\mathbf{z'}_{t-1}; \mathbf{\mu}_\theta(\mathbf{z'}_t, t, \mathbf{c}), \Sigma_\theta(\mathbf{z'}_t, t, \mathbf{c}))
\end{equation}
where the Gaussian distribution with mean $\mathbf{\mu}_\theta$ and covariance $\Sigma_{\theta}$ is parameterized by the denoising network. $\mathbf{c}$ is the text prompt and $t$ is the timestep. The inversion process derives the estimation to find $\mathbf{z'}_{t-1}$: $\frac{1}{\sqrt{\alpha_t}} \left( \mathbf{z'}_t - \frac{1 - \alpha_t}{\sqrt{1 - \bar{\alpha}_t}} \epsilon_\theta(\mathbf{z'}_t, t, \mathbf{c}) \right) + \sigma_t \mathbf{\epsilon_t}
$, where $\epsilon_\theta(\mathbf{z'}_t, t, \mathbf{c})$ is the noise prediction, $\sigma_t$ is a pre-defined hyperparameter and $\mathbf{\epsilon_t} \sim \mathcal{N}(0,\mathbf{I})$.
The denoising process aims to estimate the added noise in the forward process and acquire the denoised latent code. Here, the reverse denoising process is organized as:
\begin{equation}
\mathbf{z'}_{t-1} = F^{-1}(\mathbf{z'}_t, t, \mathbf{c})
\end{equation}
where $F^{-1}(\cdot)$ is the denoising function and $\mathbf{z'}_{t-1}$ is the predicted latent noise at step $t-1$. As $t$ approaches $0$, the latent code $\mathbf{z}_0'$ with representational abilities can be transferred into a generated image $x'\in \mathbb{R}^{H_x\times W_x \times c_2}$ by the VAE decoder $\mathrm{\Psi}^*_{d}(\cdot)$, i.e., $x' = \mathrm{\Psi}^*_{d}(\mathbf{z}_0')$.
}
\subsection{Plug-and-play Watermarking Scheme without Retraining}
In this scheme, the watermark $w\in \mathbb{R}^{H_w\times W_w \times c_3}$ can be compressed as $\varrho_w \in \mathbb{R}^{h \times w \times c_3}$ for plugging into latent space and then decoded by a watermark decoder $D_w(\cdot)$ without retraining components or entire LDMs. However, the latent code exhibits sensitivity to disturbances, making it prone to alterations that the pre-trained VAE decoder $\mathrm{\Psi}^*_{d}(\cdot)$ cannot effectively map. Thus, the plugging process aims to reduce visual differences after watermarking: 
\begin{equation}
\label{target}
\begin{aligned}
    & \min_{\mathbf{z}_{0,w}'} \| \mathrm{\Psi}^*_d(\mathbf{z}_{0,w}') - \mathrm{\Psi}^*_d(\mathbf{z}_{0}')\| \\
    & \textit{where} \quad \mathbf{z}_{0,w}' = \mathcal{G}( \mathbf{z}_{0}', \varrho_w, \alpha)
\end{aligned}
\end{equation}

Here, $\mathcal{G}(\cdot)$ is a plugging strategy with $\alpha$ modulating its strength. Then, the watermarked image generation process can be reformulated as $x_w' = \mathrm{\Psi}^*_{d}(\mathbf{z}_{0,w}')$, where $x_w'$ is the watermarked image. The watermark recovery process is calculated as $w'=D_w(x_w')$.

\section{Method}
In this section, we introduce our MarkPlugger framework in both the training phase and the inference phase.

\begin{figure*}[tb]
  \centering
  \includegraphics[height=9.2cm]{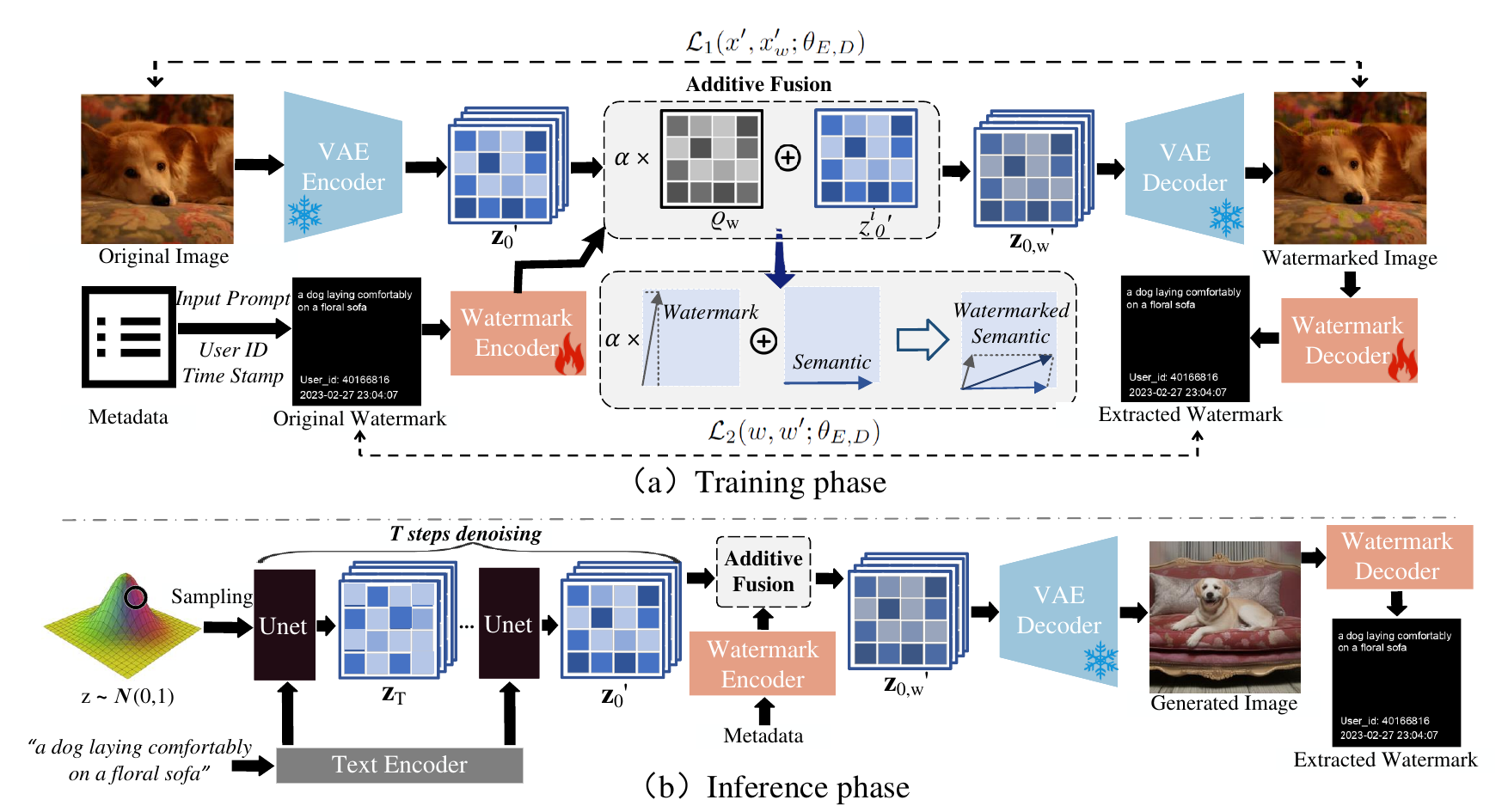}
  \caption{Overview of our proposed MarkPlugger framework for LDMs.}
  \label{fig:framework}
  \vspace{-15pt}
\end{figure*}

\subsection{Training Phase}
We describe the training process from three parts: watermark encoding, watermark decoding, and loss function. Following the watermark format outlined in~\cite{abs-2309-03815}, each watermark captures the textual prompt, the user identification, and the timestamp within pixels, which are converted into a binary image format. To enhance training efficiency while preserving latent space, we retain the VAE and exclude the UNet denoiser. Only the watermark encoder and decoder undergo gradient updates. 
The training phase of our MarkPlugger framework is shown in Fig.~\ref{fig:framework} (a).


\subsubsection{Watermark encoding}Leveraging the compressive representational capabilities of the pre-trained VAE encoder $\mathrm{\Psi}^*_{e}(\cdot)$ in the latent space, we explore the embedding of the watermark signal within the latent space. The latent code $\mathbf{z}_0'$ captures a low-level representation of the generated image, making it an ideal candidate for watermarking. To align with the representation of $\mathbf{z}_0'$, we design a watermark encoder $\mathrm{\Theta_{enc}}(\cdot)$ that utilizes a deep convolutional network (CNN) followed by a UNet. The watermark $w$ is compressed into a latent representation $\varrho_w \in \mathbb{R}^{h \times w \times c_3}$ via the encoding process: $\varrho_w = \mathrm{\Theta_{enc}}(w)$. Then, to reduce the disruption on the latent code by watermark insertion, we allocate only one channel of the latent code to carry the compressed watermark signal.  \update{Specifically, the plugging strategy $\mathcal{G}(\cdot)$ for the watermarked latent code of the $i$-th channel ${z^i_{0,w}}'$ is calculated as follows:}
\begin{equation}
\label{eq:embed_watermark}
{z^{i}_{0,w}}' = 
\begin{cases} 
{z_0^i}' + \alpha \cdot \varrho_w &  i = \kappa \\
{z_0^i}' & i \neq \kappa
\end{cases}
\end{equation}
where $\kappa$ is the designated watermarked channel. The watermarked latent code can be calculated as: $\mathbf{z}_{0,w}' = \{{z^i_{0,w}}' \cdot \beta_i\}_{i=1}^{c_1}$. The coefficient $\beta_i$ is the Kronecker delta function~\cite{jayavel2023improved}, denoted as $\beta_i=[\delta(1-i),\delta(2-i),...,\delta(c_1-i)]^T$, and $i$ is the channel number. 
\update{As shown in Eq. \ref{eq:embed_watermark}, only the selected channel $\kappa$ carries the watermark information, which aims to maintain the quality of images by limiting the perturbation on the latent space.  Besides only adding information on only one channel, another approach to reducing the perturbation of the original semantic is to optimize the encoding to achieve the watermark representation $\varrho_w$ is as orthogonal as possible to the semantic representation $\mathbf{z}'_0$. We try to achieve it by designing the corresponding loss. 
} 

After merging the watermark in the latent space, a pre-trained VAE decoder $\mathrm{\Psi}^*_{d}(\cdot)$ is utilized to recover the generated image $x'$ and the watermarked image $x_w'$ from the latent space, calculated as $x' = \mathrm{\Psi}^*_{d}(\mathbf{z}_0')$ and $x_w' = \mathrm{\Psi}^*_{d}(\mathbf{z}_{0,w}')$, respectively.

\subsubsection{Watermark decoding} After receiving a watermarked image, we develop a UNet-based~\cite{RonnebergerFB15} watermark decoder to effectively capture the deep features and local details of the images. Specifically, we adopt a multi-layer CNN $\mathrm{f_{CNN}}(\cdot)$ for coarse-grained watermark extraction in pixel space, and then the UNet decoder $\mathrm{\Phi_{dec}}(\cdot)$ is used for fine processing to output the watermark $w'$:
\begin{equation}
    w' = \mathrm{\Phi_{dec}}(\mathrm{f_{CNN}}(x_w'))
\end{equation}

Since the compression from pixel space to latent space by VAE can obscure hidden information, we opt to directly extract watermarks in the pixel space. By learning various meta-data patterns, the decoder can decode various watermarks. Furthermore, it cannot misdetect watermark signals.

\subsubsection{Loss function} We attempt to find a constraint that allows the UNet watermark encoder to achieve a state orthogonal to the known semantics. 

Here, we begin by defining two generic loss functions. One loss function $\mathcal{L}_1(x', x'_w; \theta_{E,D})$ measures the distortion between the original image $x'$ and the watermarked image $x'_w$. Another loss function $\mathcal{L}_2(w, w'; \theta_{E,D})$ measures the error between the decoded watermark $w'$ and the original watermark $w$. The parameters $\theta_{E,D}$ pertain to the watermark encoder and decoder. The total loss function $\mathcal{L}$ is a weighted sum of $\mathcal{L}_1$ and $\mathcal{L}_2$:
\begin{equation}
\label{eq:loss}
\mathcal{L}(w; \theta_{E,D}) = \lambda \mathcal{L}_1(x', x'_w; \theta_{E,D}) + (1 - \lambda) \mathcal{L}_2(w, w'; \theta_{E,D}) 
\end{equation}
where $\lambda$ is a weight. $\mathcal{L}_1$ aims to minimize the difference between $x'$ and $x'_w$ by ensuring that the watermark information in $x'_w$ is rendered irrelevant (i.e., towards orthogonal) to the image. Meanwhile, $\mathcal{L}_2$ guarantees the preservation and consistency of the hidden watermark information.




Two key aspects warrant attention. One is visual quality, referring to $\mathcal{L}_1$ in Eq. \ref{eq:loss}. As this watermark method is specifically designed for LDMs, the visual effect is an important property. We used the mean squared error (MSE) to quantify the loss between the general image $x'$ and the watermarked image $x_w'$. Furthermore, we use the Learned Perceptual Image Patch Similarity (LPIPS) \cite{ZhangIESW18} to compute at the feature level, using the pre-trained VGG~\cite{SimonyanZ14a} as the detection backbone. Another is watermark recoverability, referring to $\mathcal{L}_2$ in Eq. \ref{eq:loss}. The watermark is formed as a binary image, and the metadata is stored as pixels in it. In this case, we also use MSE to constrain the reconstruction quality of $w'$. Given these considerations, Eq.~\ref{eq:loss} is rewritten as follows:
\begin{equation}
\begin{split}
\label{common_loss}
   \mathcal{L} = & \underbrace{\gamma_0 \|x' , x_w’\|^2 + \gamma_1 \sum \omega_l \cdot \|\phi_{vgg}(x') , \phi_{vgg}(x_w’)\|^2 }_{\mathcal{L}_1} \\
   & + \underbrace{\gamma_2 \|w , w'\|^2}_{\mathcal{L}_2}
\end{split}
\end{equation}
where $\omega_l$ is the weight coefficient of layer $l$ of VGG and has been defined in advance. $\|\cdot,\cdot\|^2$ is the MSE function. $\gamma_0$, $\gamma_1$, and $\gamma_2$ are the loss weights, each regulating the balance between visual quality and watermark recoverability.

To improve the watermark robustness, we add the common post-processing attacks (e.g., Gaussian noise, rotation, cropping, etc.) with various intensity into the watermarked image, modeled as: $\hat{x}_w' = x_w' + \zeta_{n}^{m}$, where $\zeta_{n}^{m}$ denotes noise of type $m$ with intensity $n$. The more stringent requirement is to achieve the reconstruction of the watermark as $\hat{w}' = \mathrm{\Phi_{dec}}(\mathrm{f_{CNN}}(\hat{x}_w'))$. Overall, the loss function is updated as follows:
\begin{equation}
\label{total_loss}
   \mathcal{L}_r = \mathcal{L} + \gamma_3\|w,\hat{w}'\|^2
\end{equation}
where $\gamma_3$ is a loss weight. Empirically, $\gamma_0$ is set equal to the sum of $\gamma_2$ and $\gamma_3$.

\subsection{Inference Phase}
By removing the pre-trained VAE encoder $\mathrm{\Psi}^*_{e}(\cdot)$ and involving the pre-trained UNet denoiser $F^{-1}(\cdot)$ as Fig. \ref{fig:framework} (b), we then input noise and prompt to generate images. Unlike the training procedure, we embed the watermark into the denoised latent code $\mathbf{z}'_0$, which has undergone a multistep denoising process. This approach allows us to implement a plug-and-play watermark solution for LDMs without requiring retraining. Additionally, it can be applied across different variants of LDM.

\section{Experiments}
This section presents our experimental findings. First, we introduce the basic settings about the datasets, models, metrics in Sec.~\ref{datasets_models_metrics}, and implementation details in Sec.~\ref{implementation_details}. Second, our evaluation focuses on the model properties about watermark invisibility, watermark recoverability, image quality in Sec.~\ref{Results_without_Attacks}, model generalization in Sec.~\ref{generalization}, and watermark robustness in Sec.~\ref{robustness}. Additionally, we present the parameters analysis about watermark strength and training loss weight in Sec.~\ref{params}, and channel selection in Sec.~\ref{channels}. Finally, the visualizations further support the effectiveness of our MarkPlugger in Sec.~\ref{visualizations}.

\vspace{-3ex}

\subsection{Datasets, Models and Metrics}
\label{datasets_models_metrics}

\subsubsection{Datasets} We have constructed a dataset called COCO Subset from the commonly used dataset MS-COCO~\cite{LinMBHPRDZ14}, with $82,783$ and $10,000$ text-image pairs for training and inference, respectively. Besides, to evaluate the generalization to more examples, we then build a test set of $8,091$ images sourced from Flickr-8k\footnote{https://www.kaggle.com/datasets/adityajn105/flickr8k}.

\subsubsection{Models} To show the generalization capabilities of our MarkPlugger on LDMs, we conducted experiments on the official versions of SD, specifically v1-1, v1-4, v1-5, v2-0, v2-1~\cite{RombachBLEO22} and XL~\cite{PodellELBDMPR24}. Moreover, the prominent modifications like Taiyi~\cite{fengshenbang} and Safe Latent Diffusion~\cite{SchramowskiBDK23} are also involved. We have trained our watermark model on SD v1-1 and directly transferred it to other variants, without retraining.

\subsubsection{Metrics} Referring to the previous work~\cite{abs-2309-03815}, we inherit the evaluation system and divide the metrics into three distinct perspectives: watermark invisibility, watermark recoverability, and image quality. For watermark invisibility, we use the Peak Signal-to-Noise Ratio (PSNR)~\cite{DingMCL22}, Structural Similarity (SSIM)~\cite{DingMCL22}, and Learned Perceptual Image Patch Similarity (LPIPS)~\cite{ZhangIESW18} to measure the imperceptibility of hidden watermarks at the pixel level and feature level. For watermark recoverability, we use Normalized Correlation (NC)~\cite{TancikMN20} and Character Account (CA)~\cite{abs-2309-03815} based on edit distance to measure the robustness of the watermarks at the pixel level and character level. Since different examples have different prompt lengths, we develop a metric to show the ratio of edited characters compared to the total characters $n_i$ of the $i$-$th$ prompt, called Character Edit Ratio (CER), which is calculated as $CER = \{\sum_{i=1}^{N} {CA_{i}}/n_i\}/N \times 100\%$, where $N$ is the number of prompts. For image quality, we use $\Delta$ FID to calculate the difference in Fréchet Inception Distance (FID)~\cite{HeuselRUNH17} between the original and the watermarked images. The proportion of $\Delta$ FID in original FID is denoted as $p_{\Delta \text{FID}}$.

\begin{table*}[!t]
  \footnotesize 
  \renewcommand\arraystretch{1.2}
  \centering
  \setlength{\tabcolsep}{2.65mm}{
		\caption{ Evaluation about image invisibility, watermark recoverability and image quality of our framework applicable to eight variants of LDM on two datasets. * denotes that the watermark model has been trained. $^\dagger$ denotes that the watermark model is directly ported from SD-MarkPlugger v1-1* on COCO Subset without retraining.}
		\label{tb:wo_attack}
		\vspace{2ex}
		\begin{tabular}{cccccccccccc}
			\toprule
			\multirow{2}{*}{Datasets}&\multirow{2}{*}{Models}
			&\multicolumn{3}{c}{Watermark Invisibility}&\multicolumn{3}{c}{Watermark Recoverability}
   &\multicolumn{2}{c}{Image Quality}\\
			\cline{3-10}  &&PSNR(dB)$\uparrow$&SSIM(\%)$\uparrow$&LPIPS$\downarrow$&NC(\%)$\uparrow$&CA$\downarrow$&CER(\%)$\downarrow$&$\Delta$ FID$\downarrow$& $p_{\Delta \text{FID}}(\%)\downarrow$\\
			\midrule 
            \multirow{11}{*}{\rotatebox{90}{\textbf{COCO Subset}}}
            &HiDDeN~\cite{ZhuKJF18}	&$32.19$&$92.33$&$\textbf{3.79}$\textbf{e}-$\textbf{02}$&-&-&- &$-1.30$&$4.88$\\
            &SSL watermark~\cite{FernandezSFJD22}	&$31.24$&$90.29$&$8.35$\text{e}-$02$&-&-&- &$-0.98$&$3.68$\\
            &Stable Signature~\cite{FernandezCJDF23}	&$31.06$&$90.67$&$4.99$\text{e}-$02$&-&-&- &$-\textbf{0.54}$&$\textbf{2.22}$\\
			&\cellcolor{gray!20}
   \textbf{SD-MarkPlugger v1-1*}~\cite{RombachBLEO22}&\cellcolor{gray!20}$37.04$&\cellcolor{gray!20}$94.35$& $\cellcolor{gray!20}$4.21$\text{e}$-$02$&\cellcolor{gray!20}$96.15$&\cellcolor{gray!20}$13.31$&\cellcolor{gray!20}$14.20$&\cellcolor{gray!20}$-1.35$&\cellcolor{gray!20}$5.07$ \\
            &SD-MarkPlugger v1-4$^\dagger$~\cite{RombachBLEO22}
			&$36.93$ &$94.10$& $4.32$\text{e}-$02$&$\textbf{97.14}$&$11.97$&$13.32$&$-1.33$&$5.03$ \\
           &SD-MarkPlugger v1-5$^\dagger$~\cite{RombachBLEO22}
			&$36.97$& $94.46$&$4.24$\text{e}-$02$&$96.78$&$11.85$ &$13.33$&$-1.50$&$5.76$ \\
   &SD-MarkPlugger v2-0$^\dagger$~\cite{RombachBLEO22}
			&$\textbf{37.19}$& $94.41$&$4.38$\text{e}-$02$&$97.09$&$12.06$&$15.13$&$-1.52$&$5.27$ \\
   &SD-MarkPlugger v2-1$^\dagger$~\cite{RombachBLEO22}
			&$37.02$& $94.30$&$4.41$\text{e}-$02$&$97.01$&$\textbf{11.71}$&$\textbf{12.74}$&$-1.50$&$4.87$ \\
   &SD-MarkPlugger XL$^\dagger$~\cite{PodellELBDMPR24}
			&$36.83$& $94.02$&$4.23$\text{e}-$02$&$96.72$&$12.15$&$15.29$&$-1.44$&$4.94$ \\
   &Taiyi-MarkPlugger$^\dagger$~\cite{fengshenbang}
			&$36.75$& $\textbf{94.51}$&$4.33$\text{e}-$02$&$96.82$&$12.12$&$13.49$&$-1.49$&$4.62$ \\ 
   &Safe LDM-MarkPlugger$^\dagger$~\cite{SchramowskiBDK23}
			&$36.89$& $94.33$&$4.28$\text{e}-$02$&$96.91$&$11.99$&$15.13$&$-1.55$&$4.84
   $ \\ 
            \midrule
            \multirow{11}{*}{\rotatebox{90}{\textbf{Flikr-8K}}}

            &HiDDeN~\cite{ZhuKJF18}	&$32.03$&$92.12$&$\textbf{4.05}$\textbf{e}-$\textbf{02}$&-&-&- &$-1.79$&$4.58$\\
            &SSL Watermark~\cite{FernandezSFJD22}	&$31.17$&$90.44$&$7.93$\text{e}-$02$&-&-&- &$-1.59$&$4.08$\\
            
            &Stable Signature~\cite{FernandezCJDF23}
			&$30.84$&$90.28$&$5.06$\text{e}-$02$&-&-&- &$-\textbf{0.50}$&$\textbf{1.51}$\\
			&SD-MarkPlugger v1-1$^\dagger$~\cite{RombachBLEO22}
			&$36.77$&$93.74$& $4.46$\text{e}-$02$&$96.15$&$13.31$&$14.20$&$-2.20$&$5.64$ \\
            &SD-MarkPlugger v1-4$^\dagger$~\cite{RombachBLEO22}
			&$36.64$& $93.34$& $4.51$\text{e}-$02$&$96.67$&$13.67$&$14.83$&$-1.97$&$5.14$\\
           &SD-MarkPlugger v1-5$^\dagger$~\cite{RombachBLEO22}
			&$36.72$& $\textbf{93.82}$&$4.36$\text{e}-$02$&$96.44$&$14.40$&$15.13$ &$-2.22$&$5.79$\\
   &SD-MarkPlugger v2-0$^\dagger$~\cite{RombachBLEO22}
			&$36.47$& $93.80$&$4.26$\text{e}-$02$&$\textbf{96.73}$&$13.95$&$14.91$&$-2.01$&$5.33$ \\
   &SD-MarkPlugger v2-1$^\dagger$~\cite{RombachBLEO22}
			&$\textbf{36.94}$& $93.79$&$4.35$\text{e}-$02$&$96.62$&$\textbf{13.20}$&$\textbf{13.97}$&$-1.86$&$4.72$ \\
   &SD-MarkPlugger XL$^\dagger$~\cite{PodellELBDMPR24}
			&$36.53$& $93.71$&$4.33$\text{e}-$02$&$96.23$&$14.78$&$15.42$&$-1.91$&$4.55$ \\
   &Taiyi-MarkPlugger$^\dagger$~\cite{fengshenbang}
			&$36.72$& $93.54$&$4.41$\text{e}-$02$&$96.19$&$14.03$&$15.10$&$-2.14$&$4.59$ \\ 
   &Safe LDM-MarkPlugger$^\dagger$~\cite{SchramowskiBDK23}
			&$36.69$& $93.46$&$4.29$\text{e}-$02$&$96.50$&$13.64$&$14.72$&$-2.03$&$4.70$ \\ 
			\bottomrule
		\end{tabular}
	}
 
	\vspace{-4ex}
\end{table*}

\vspace{-2ex}
\subsection{Implementation Details}
\label{implementation_details}
The size of the generated image is $512\times512$. The SD sampler and pipeline are DDPMScheduler and StableDiffusionPipeline. The number of inference steps is set to $30$. The size of the latent code is originally set to $64\times64\times4$. The size of the compressed watermark representation is embedded in $64\times64\times1$. The watermark signal is attached to the last channel of the latent code, that is, $\kappa$ is $3$. $\gamma_0$, $\gamma_1$, $\gamma_2$, and $\gamma_3$ are empirically set at $2$, $0.2$, $1$, and $1$. The binary watermark is defined as a single channel with a size of $256\times256$. The input prompt, user ID (sampling randomly from the distribution $\mathbb{U}(0,9)$) and time stamps are displayed in pixels. In our main experiment, we train our watermark model with SD v1-1 on the COCO Subset train split only for one epoch. Experiments are conducted on an NVIDIA A800 GPU, but our method is also feasible on the lower-memory platforms.

\vspace{-2.7ex}
\subsection{Performances without Attacks}
\label{Results_without_Attacks}
As discussed in Sec.~\ref{datasets_models_metrics}, we measure the effectiveness of our MarkPlugger framework in the plug-and-play scheme without retraining from three perspectives: watermark invisibility, watermark recoverability, and image quality. In this comparison, we benchmark our approach on the official versions and frequently used modifications of LDM. As shown in Tab. \ref{tb:wo_attack}, we adopt our framework on eight variants across two datasets to show the above superiority. Besides comparing MarkPlugger with two post-embedding methods~\cite{ZhuKJF18,FernandezSFJD22} on the images generated by SD v1-1, we also consider open-source Stable Signature~\cite{FernandezCJDF23} as our baseline. Since it fixes a binary sequence as the watermark, we mainly focus on its watermark invisibility and image quality. Another popular method, ENDE~\cite{Xiong0F023} lacks accessible code, and we will only consider its computational efficiency and visualization result later.

\subsubsection{Watermark Invisibility}At the pixel level, we evaluate the outstanding invisibility of our watermark using PSNR and SSIM metrics. Across all model versions and datasets tested, our PSNR consistently registers at no less than $36.47$ dB, comfortably above the $30$ dB benchmark for high-quality visual effects~\cite{yako2023video}. In comparisons between original and watermarked images, our SSIM scores surpass those achieved by numerous conventional methods~\cite{ZhuKJF18,FernandezSFJD22}. Shifting to the feature level, we employ LPIPS to further assess the watermark invisibility. Our method achieves lower LPIPS scores compared to those obtained by Stable Signature~\cite{FernandezCJDF23}. The hidden watermarks in textures are less easily captured by human perceptual abilities.

\subsubsection{Watermark Recoverability}At the pixel level, NC demonstrates more than 96\% pixel alignment between the original and extracted watermarks, indicating high fidelity. Furthermore, we calculate the CER to compute the proportion of modified characters in the ground truth, with approximately $12\% \sim 15\%$ of characters diminishing recognizability.

\subsubsection{Image Quality}After adding watermarks to the images, we observe that performance fluctuates by approximately $4\% \sim 5\%$ across different versions. Stable Signature demonstrates slightly greater stability, as it only embeds one fixed watermark, sacrificing the flexibility of information embedding.


\subsection{Watermark Model Generalization}
\label{generalization}
We have trained our MarkPlugger on SD v1-1 for just one epoch to obtain all the results, as demonstrated in Tab.~\ref{tb:wo_attack}. Upon transferring to other variants of LDM without training any component, no significant disruption is observed across all metrics. The good generalization capabilities evidence the adaptability of our watermark model to version iterations of LDM. We analyze the origins of its generation as follows. 

\subsubsection{Adaptability to the frozen VAE decoder}With the universal representation capability of the watermark encoder, the watermark can adapt to the latent space. Moreover, the watermark signal induces only a minimal offset to the original latent code, ensuring compatibility with frozen VAE decoders across different versions. Therefore, the watermark can also be restored by various VAE decoders and hidden in pixels.

\subsubsection{Perception ability of our watermark decoder}Despite the utilization of diverse training data across various iterations of LDM, the foundational mechanism of watermarked image generation remains consistent. Specifically, it involves the reconstruction of the watermarked image from a latent space. Our proposed methodology, especially a UNet-based watermark decoder, is adept at capturing the latent watermark signal after the mapping by the VAE decoder. Consequently, it is capable of facilitating watermark decoding across variants of LDM.


\begin{figure}[tb]
  \centering
  \includegraphics[height=4.3cm]{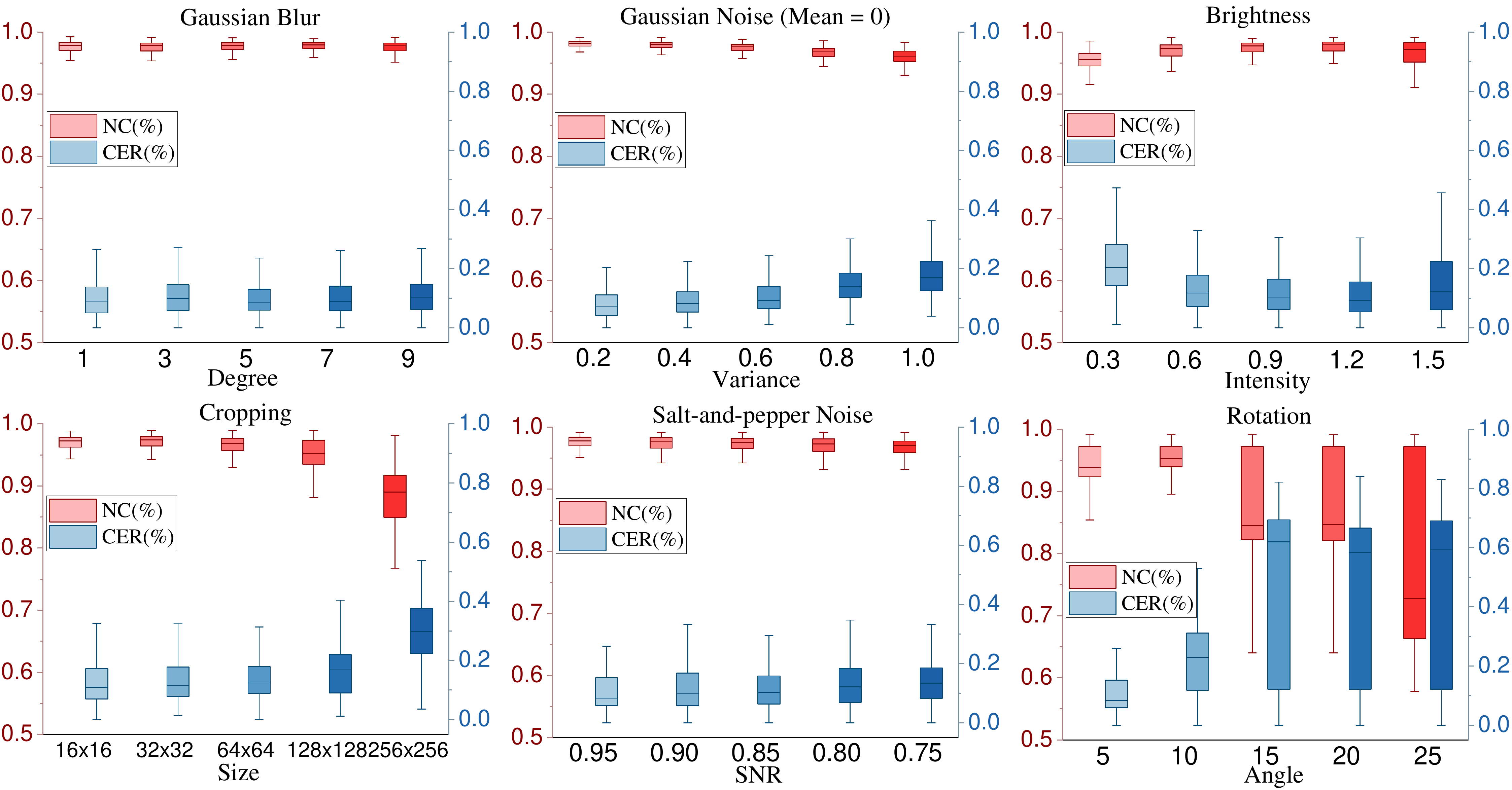}
  \caption{Robustness under various attacks with different perturbation strengths on COCO. A darker hue indicates a more potent attack.}
  \label{fig:robust}
  \vspace{-10pt}
\end{figure}

\begin{figure}[tb]
  \centering
  \includegraphics[height=4.3cm]{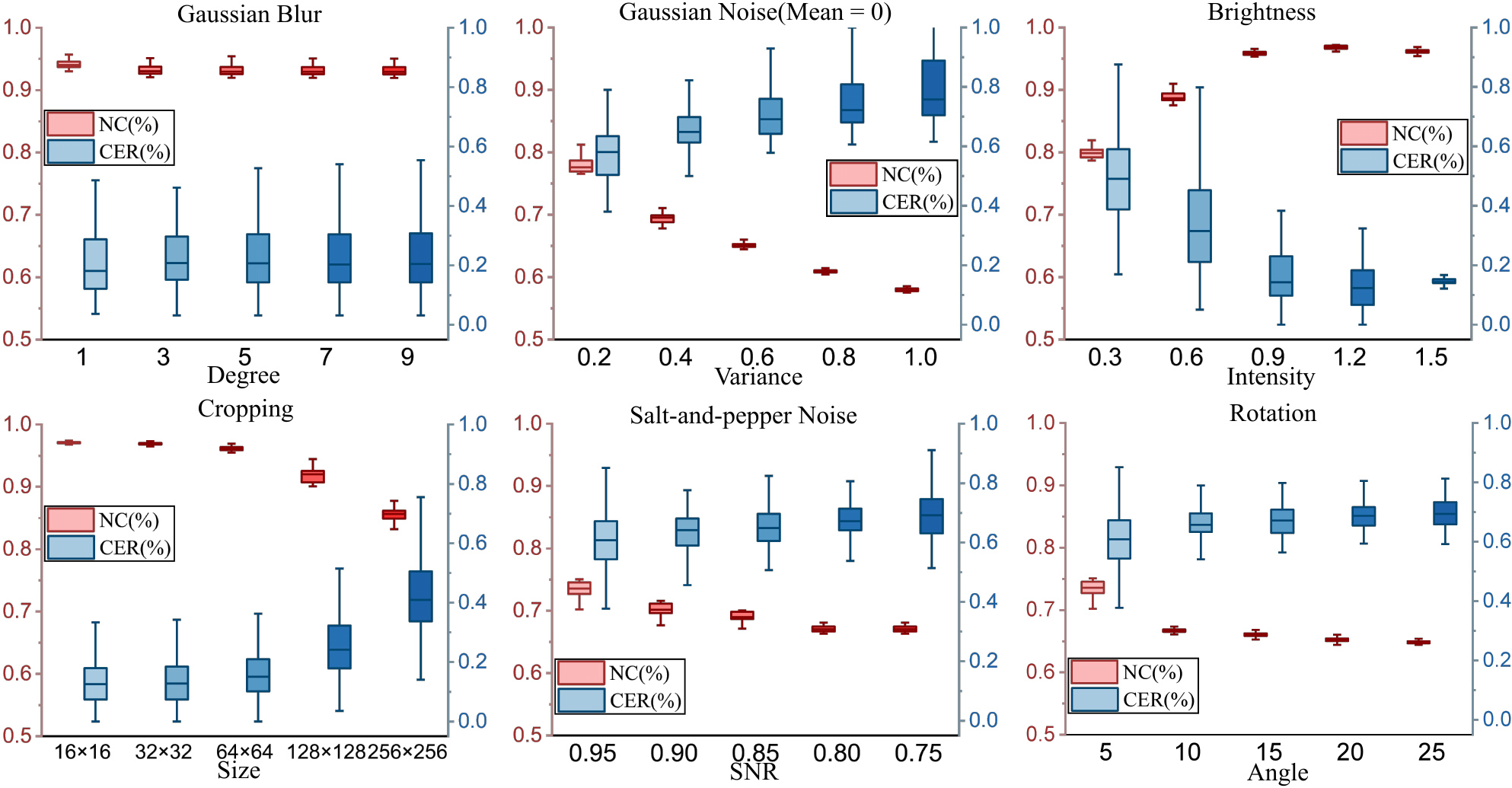}
  \caption{\update{Robustness under various attacks with different perturbation strengths on Flikr-8K with the same setting as Fig.~\ref{fig:robust}.}}
  \label{fig:robust_FL}
  \vspace{-10pt}
\end{figure}

\subsection{Robustness against Attacks}
\label{robustness}
We validate the watermark robustness of our MarkPlugger on SD v1-5 of COCO Subset. As shown in Fig.~\ref{fig:robust}, we select NC and CER to quantify the watermark performance from the perspective of pixel-level and character-level. For Gaussian blur, the watermark is almost unaffected. When encountering Gaussian noise with $\mu = 0$, the watermark is minimally violated as the intensity of the noise variance increases. The brightness slightly affects the distribution of the generated watermarks while intensity $= 1$ denotes the original state. It is also acceptable for cropping because the $256 \times 256$ cropping size allows the mean NC to still approach 90$\%$. Our watermark can also resist salt-and-pepper noise. Relatively, the higher angle rotation can destruct our watermark to a greater extent. In conclusion, the watermarks show good robustness to attacks on COCO Subset.

\update{In addition, we also evaluate the watermark robustness on Fliker-8K dataset to check the cross-dataset performance.  As shown in Fig.~\ref{fig:robust_FL}, the watermark recovery performance maintains almost good under Gaussian blur, brightness and cropping. However, for the rest of three attacks, the robustness is not as promising as COCO where the watermark encoder and decoder are trained. To sum up, the one epoch training watermark model has some robustness on different datasets, but how to make its robustness more generalizable remains to be investigated. }

\begin{figure}[tb]
  \centering  \includegraphics[height=2.3cm]{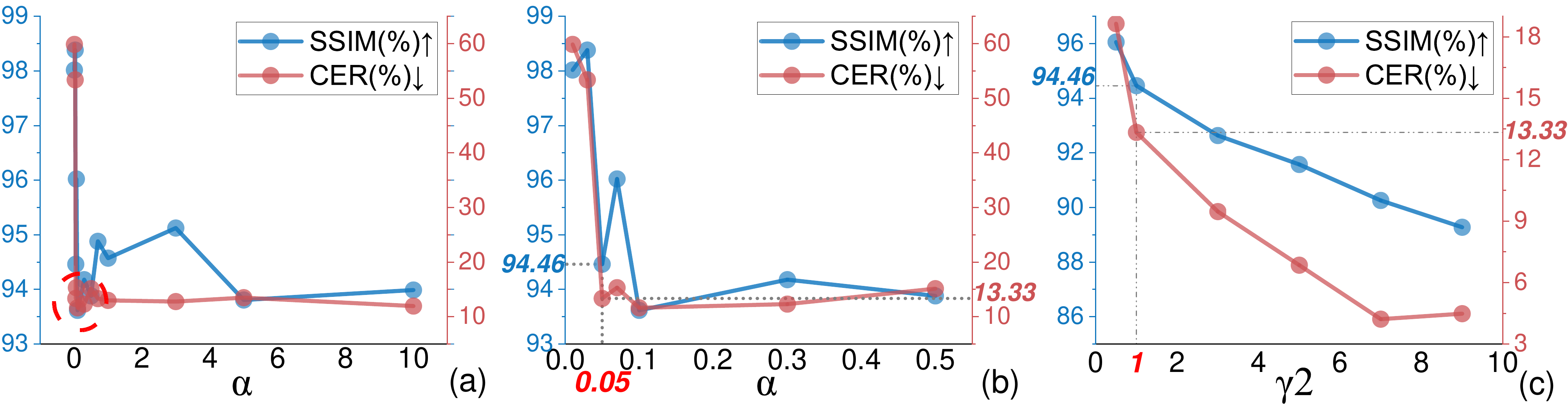}
  \vspace{-2ex}
  \caption{Parameter analysis on COCO.}
  \label{fig:param}
  \vspace{-10pt}
\end{figure}

\begin{figure}[tb]

  \centering  \includegraphics[height=2.3cm]{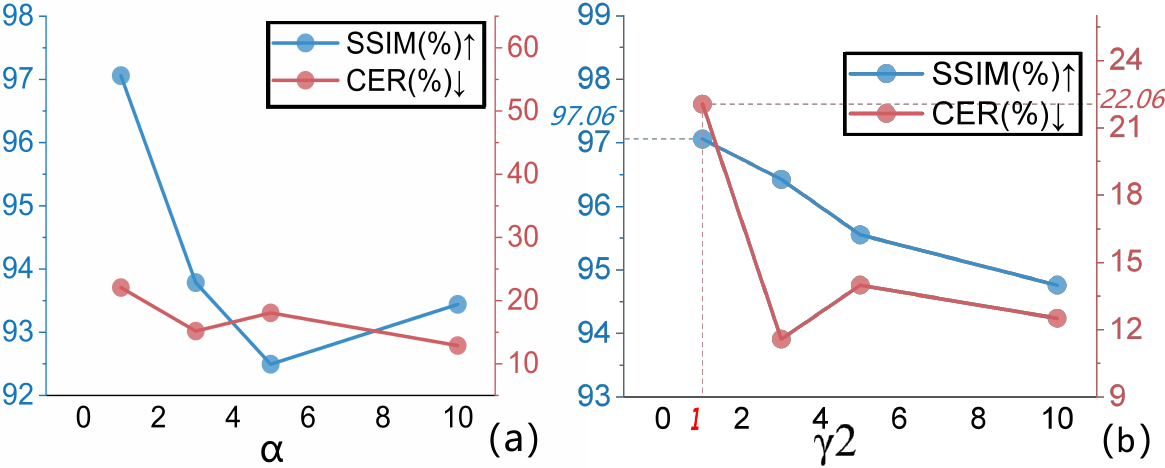}
  \vspace{-2ex}
  \caption{\update{Parameter analysis on Flikr-8K.}}
  \label{fig:param-FL}
  \vspace{-10pt}
\end{figure}

\subsection{Parameters Analysis}
\label{params}
In our framework, two parameters need to be further discussed, i.e., the coefficient of watermark signal strength $\alpha$ and loss weight $\gamma_2$. Both serve to balance watermark invisibility and watermark recoverability. We list the curves of SSIM and CER in Figs.~\ref{fig:param} and \ref{fig:param-FL} to show their negative correlation. We sample a range of $\alpha$ from $0$ to $10$ in Fig.~\ref{fig:param}(a).  Fig.~\ref{fig:param}(a) reveals that the SSIM curve decreases and the CER curve increases until $\alpha$ reaches a value. More clearly, we enlarge the red circle in Fig.~\ref{fig:param}(b). After $\alpha$ exceeds $0.1$, these curves almost stabilize. We find $\alpha = 0.05$ as the equilibrium state and manually set it in the main experiments. Furthermore, Fig.~\ref{fig:param}(c) shows that loss of weight $\gamma_2$ has a more severe impact on balance. In addition, it enables the regulation of the negative correlation between watermark invisibility and watermark recoverability. Empirically, we have set $\gamma_2$ to $1$ by default.

\begin{table}[!t]
	\scriptsize
    \renewcommand\arraystretch{1.4}
	\centering
	\scriptsize
	\setlength{\tabcolsep}{0.27mm}{
		\caption{ Results of embedding the watermark into different channels in latent space.}
		\label{tb:channel_selection}
		\vspace{-2ex}
        
		\begin{tabular}{ccccccccc}
			\toprule
			\multirow{2}{*}{Num}
			&\multicolumn{3}{c}{Watermark Invisibility}&\multicolumn{3}{c}{Watermark Recoverability}&\multicolumn{2}{c}{Image Quality}\\
			\cline{2-9}  &PSNR(dB)$\uparrow$&SSIM(\%)$\uparrow$&LPIPS$\downarrow$&NC(\%)$\uparrow$&CA$\downarrow$&CER(\%)$\downarrow$&$\Delta$ FID$\downarrow$& $p_{\Delta \text{FID}}\downarrow$\\
			\midrule 
			Channel0 &$\textbf{37.18}$&$\textbf{94.62}$& $4.25$\text{e}-$02$&$95.97$&$13.04$&$14.34$&$-2.33$&$8.95$ \\
            Channel1
			&$36.61$&$94.04$& $4.22$\text{e}-$02$&$\textbf{96.88}$&$\textbf{11.45}$&$\textbf{13.26}$&$-1.75$&$6.72$ \\
           Channel2
			&$36.88$& $94.51$&$\textbf{4.19}$$\textbf{e}$-$\textbf{02}$&$96.42$&$12.37$ &$14.01$&$-1.83$&$7.03$ \\
            Channel3
			&$36.97$& $94.46$&$4.24$\text{e}-$02$&$96.78$&$11.85$ &$13.33$&$-\textbf{1.50}$&$\textbf{5.76}$ \\
			\bottomrule
		\end{tabular}
	}
	\vspace{-4ex}
\end{table}

\vspace{-2ex}
\subsection{Watermarking Different Channels in Latent Space}
\label{channels}
As indicated in Eq. \ref{eq:embed_watermark}, the watermark signal is incorporated by attaching it to one channel. In the latent space, the latent code of the image is structured as $64 \times 64 \times 4$. Consequently, we endeavor to embed the watermark in each channel to evaluate the impact of watermark carrier selection on the results, as detailed in the Tab.~\ref{tb:channel_selection}. We utilize SD v1-5 to perform tests on the COCO Subset. Remarkably, the performance exhibits slight fluctuations within an acceptable range for all metrics. In conclusion, our framework demonstrates the feasibility of embedding watermarks across all channels. We have selected channel $3$ as the standard setting. 

\begin{table}[!t]
	\scriptsize
    \renewcommand\arraystretch{1.4}
	\centering
	\scriptsize
	\setlength{\tabcolsep}{6.75mm}{    
 \caption{Flops computation for components requiring training.}
 \label{tb:flops}
    {\footnotesize
    \begin{tabular}{ccc}
        \toprule
        Models
			&FLOPs(G)
			&Params(M)
             \\
            \midrule
            Stable Signature \cite{FernandezCJDF23}
			& $2767.15$ & $50.84$\\
            \midrule
            ENDE \cite{Xiong0F023}
			& $3254.42$  &$67.13$\\
            \midrule
            MarkPlugger (Ours)
			&$\textbf{9.72}$ & $\textbf{0.97}$\\
\bottomrule
    \end{tabular}
    }}
    \vspace{-8pt}
\end{table}

\begin{figure}[t]
  \centering
  \includegraphics[height=4.cm]{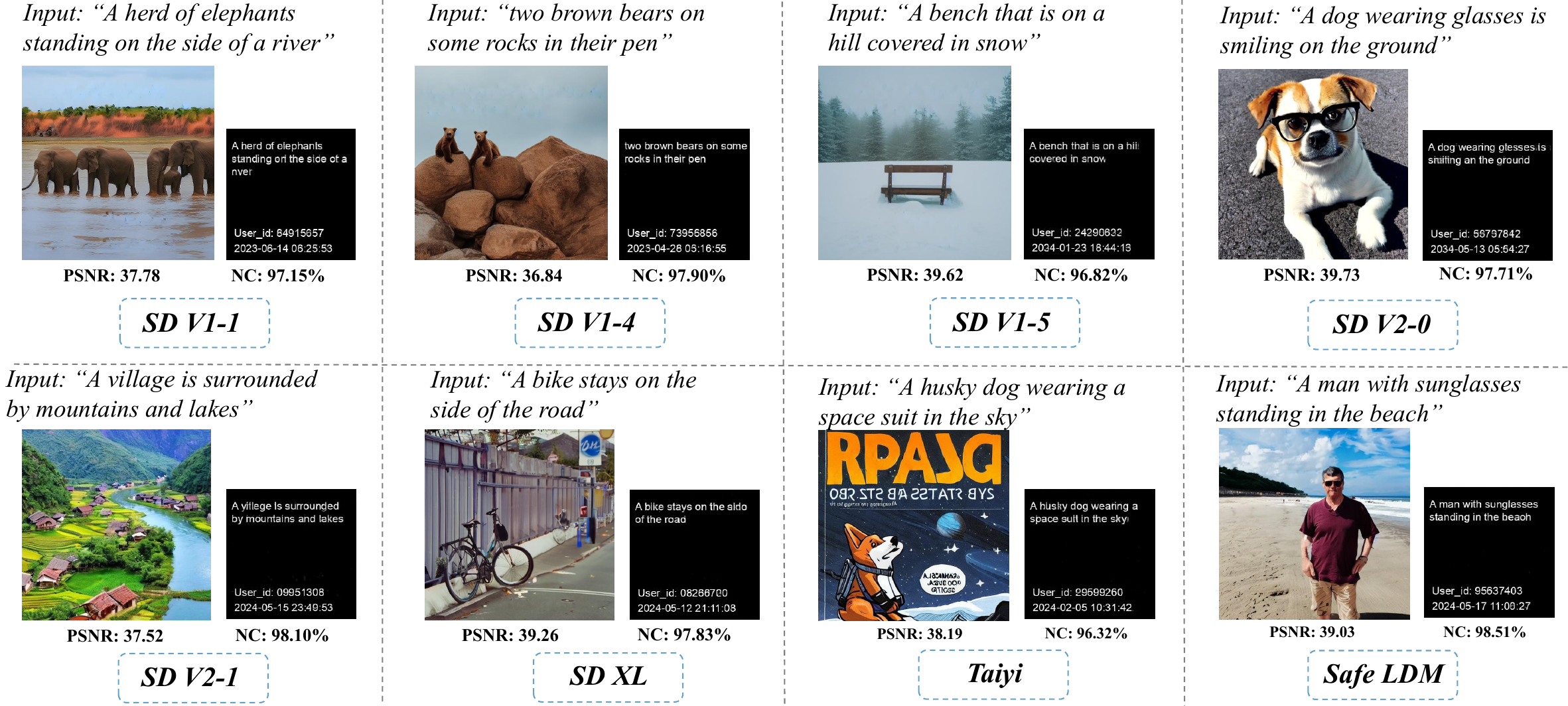}
  \caption{Visual results of MarkPlugger on different variants of LDM.}
  \label{fig:official}
  \vspace{-10pt}
\end{figure}

\begin{figure}[t]
  \centering  \includegraphics[height=2.785cm]{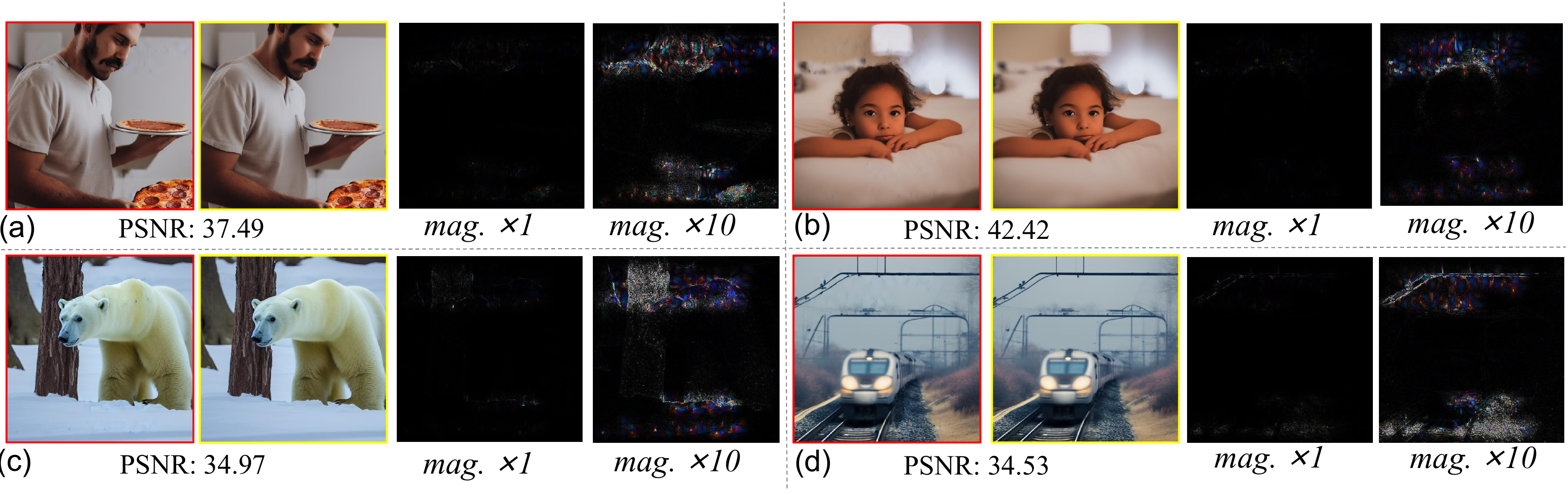}\
  \caption{The pixel difference at magnifications of one and ten times between the watermarked images (highlighted with a red border) and original images (highlighted with a yellow border).}
  \label{fig:difference}
  \vspace{-10pt}
\end{figure}

\vspace{-6pt}
\subsection{FLOPs Computation}
\label{flops}
Our MarkPlugger focuses on lightweight and high efficiency. As shown in Tab.~\ref{tb:flops}, we statistically quantify the FLOPs~\cite{PeeblesX23} for the trainable parameters of our framework, in comparison to recent advancements such as Stable Signature \cite{FernandezCJDF23} and ENDE \cite{Xiong0F023}, both of which retrain the LDM. For ENDE, we calculate its FLOPs and parameters based on the parameters provided in the paper~\cite{Xiong0F023}. Our framework achieves significantly lower FLOPs across orders of magnitude, highlighting its computational efficiency compared to these methods. Additionally, it operates with a minimal parameter set of just $0.97$ million, which is nearly $1/70$ of ENDE's parameters. It still facilitates the embedding of diverse watermarks in LDM variants.

\vspace{-3ex}
\subsection{Visualizations}
\label{visualizations}


\subsubsection{Application on official versions and modified variants} We apply our MarkPlugger framework to $8$ LDM versions and variants, as displayed in Fig.~\ref{fig:official}. The results demonstrate a well-balanced trade-off between image quality and watermark recoverability, maintaining a high standard. 

\begin{figure}[t]
  \centering
  \includegraphics[height=4.5cm]{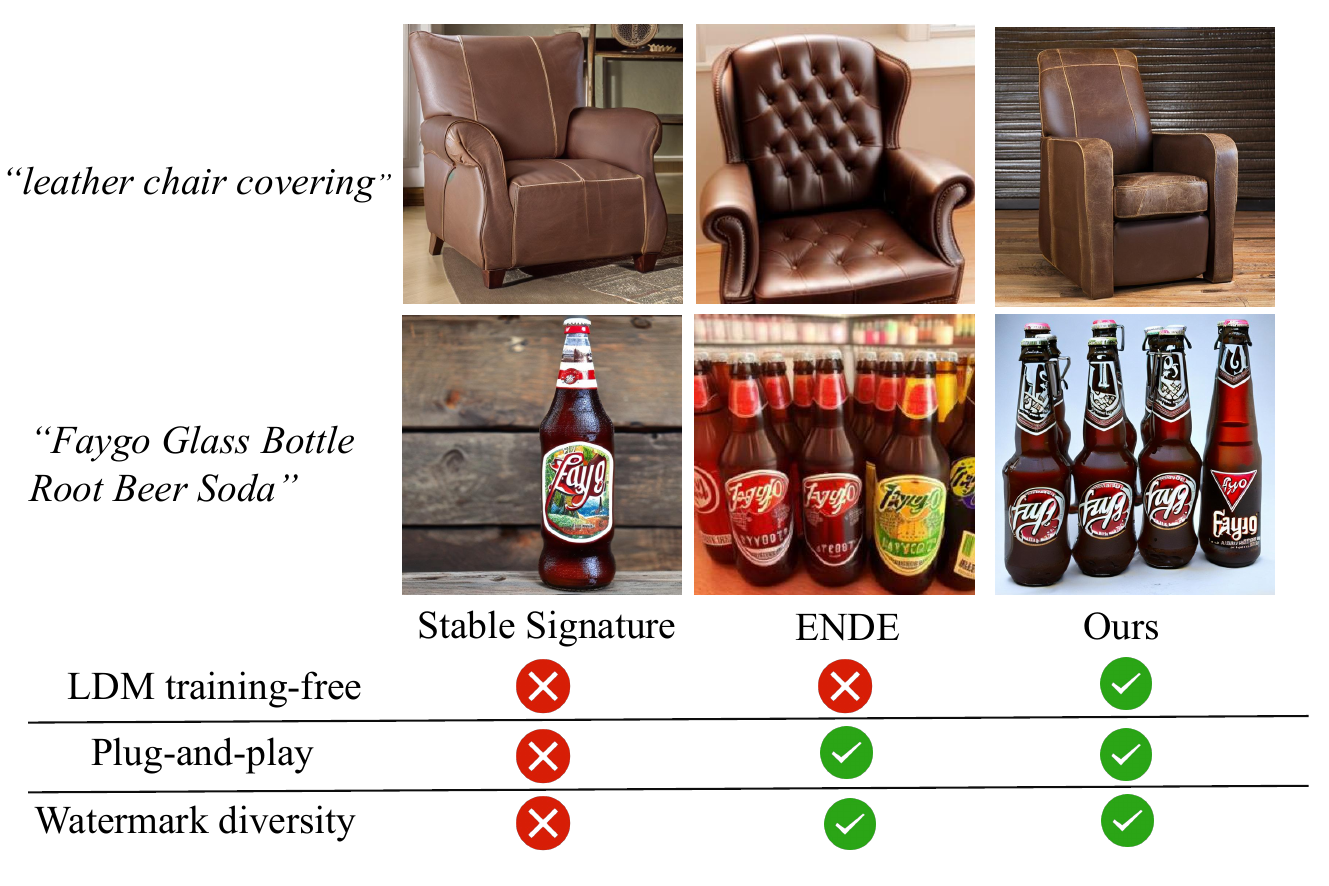}
  \setlength{\abovecaptionskip}{-0.1cm}
      \caption{Comparison of watermarked images generated by our framework with two other methods for LDM. The green and red checkmarks denotes the presence or absence of properties.}
  \vspace{-10pt}
  \label{fig:comparison}
\end{figure}

\begin{figure}[t]
  \centering
  \includegraphics[height=1.6cm]{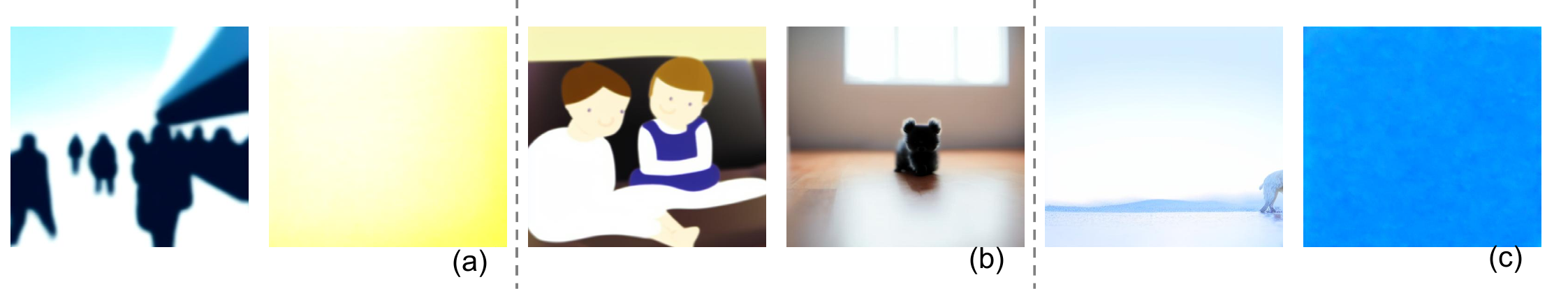}
  \caption{The generated images from different positions of watermark plugging. }
  \label{fig:embedding_position}
  \vspace{-10pt}
\end{figure}

\begin{figure}[t]
  \centering
  \includegraphics[height=2.7cm]{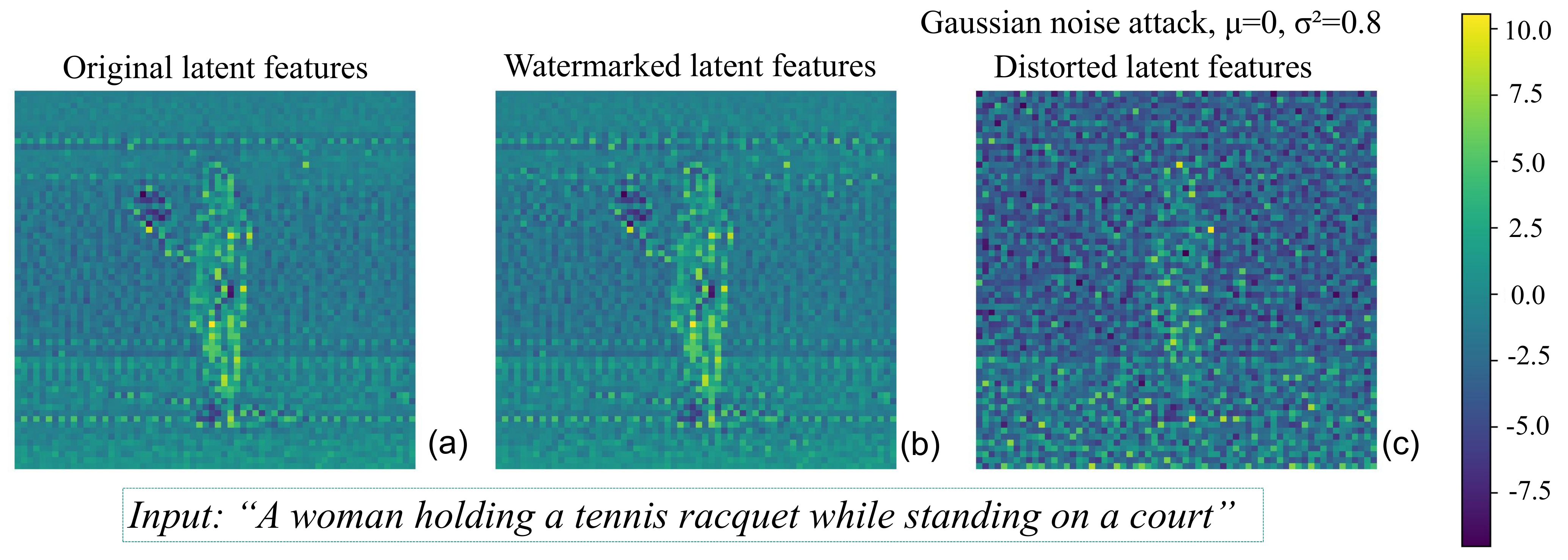}
  \setlength{\abovecaptionskip}{-0.1cm}
      \caption{Visualizations of three types of latent features. The watermarked latent features and the distorted latent features carry watermarks.}
  \vspace{-16pt}
  \label{fig:latent}
\end{figure}

\subsubsection{Image difference}Fig.~\ref{fig:difference} presents four illustrative examples, along with their computed PSNR, highlighting the differences between the watermarked images and the original images. The divergence between the two images is subtle, making it difficult to discern by human observation. Upon differentiation, a sparse distribution of non-zero pixel values emerge, highlighting the regions affected by watermark embedding.

\subsubsection{Comparison with other watermark methods}Fig.~\ref{fig:comparison} shows that our MarkPlugger framework achieves considerable visual effect on watermarked images with diverse watermarks, compared to the training-based method called ENDE ~\cite{Xiong0F023} and the model-customized method called Stable Signature~\cite{FernandezCJDF23}.

\subsubsection{The position of watermark embedding}To figure out the most suitable position in the LDM generation phrases to embed the watermark, we list our attempts in Fig.~\ref{fig:embedding_position}. First, we embed the watermark into the initial Gaussian noise sampling, which causes the LDM to produce a collapsed image, as shown in Fig.~\ref{fig:embedding_position} (a). We guess that the watermark signal disrupts the distribution of the late code. Therefore, we then use the Yeo-Johnson transformation \cite{weisberg2001yeo} to increase the normality of the latent code in Fig.~\ref{fig:embedding_position} (b). Although the visual effect has been improved, it still cannot generate normal images. Finally, we embed the watermark into latent code during denoising several dozen steps. The model also collapses, as shown in Fig.~\ref{fig:embedding_position} (c). All failure attempts cannot resolve any watermark information. Finally, we successfully embed the watermark after denoising.

\subsubsection{Features in latent space}
As suggested, we depict the latent features in Fig.~\ref{fig:latent}. Compared to the original latent features, the watermark signal produces slight disturbance in the watermarked latent features. Our watermark decoder can capture the subtle watermark signal from both watermarked and distorted features.

\section{Conclusion}
In this study, we propose MarkPlugger, a generalizable plug-and-play watermark framework for LDMs without retraining. Unlike previous training-based and model-customized watermark approaches, our method compresses the watermark signal within latent space and embeds it with the latent code, efficiently achieving the generation of watermarked images without retraining components or the entire LDM. Moreover, we reach a balance between image quality and watermark recovery. It also shows good robustness against attacks. Additionally, our framework exhibits exceptional generalization capabilities and can be easily transferred to multiple variants of LDM.

\bibliography{egbib}


\end{document}